\documentclass[lettersize,journal]{IEEEtran}
\usepackage{amsmath,amsfonts}
\usepackage{algorithmic}
\usepackage{algorithm}
\usepackage{array}
\usepackage[caption=false,font=normalsize,labelfont=sf,textfont=sf]{subfig}
\usepackage{textcomp}
\usepackage{stfloats}
\usepackage{url}
\usepackage{verbatim}
\usepackage{graphicx}
\usepackage{cite}
\usepackage{multirow}
\usepackage{comment}
\usepackage{colortbl}
\usepackage{caption}
\usepackage{hyperref}
\usepackage{booktabs}
\usepackage{colortbl}
\usepackage{amsmath,amsfonts}
\usepackage{algorithmic}
\usepackage{algorithm}
\usepackage{array}
\usepackage{textcomp}
\usepackage{stfloats}
\usepackage{url}
\usepackage{verbatim}
\usepackage{graphicx}
\usepackage{cite}
\usepackage{subcaption}
\begin{document}

\title{GaussianEmoTalker: Real-Time Emotional Talking Head Synthesis with Audio-Driven and Blendshape-Based 3D Gaussian Splatting}



\twocolumn

\author{Haijie Yang, Zhenyu Zhang, Yixuan Dong, Jianjun Qian~\IEEEmembership{Member,~IEEE,} Jian Yang
\thanks{This work was supported by the NSFC under Grant Nos. U24A20330, 62361166670, 62376121, 62176124 and Postgraduate Research \& Practice Innovation Program of Jiangsu Province. 

Haijie Yang, Jianjun Qian, and Jian Yang are with PCA Lab, Key Lab of Intelligent Perception and Systems for High-Dimensional Information of Ministry of Education, School of Computer Science and Engineering, Nanjing University of Science and Technology, Nanjing 210094, China (e-mail: yanghaijie@njust.edu.cn; csjqian@njust.edu.cn; csjyang@njust.edu.cn). 

Yixuan Dong, Tsientang Institute for Advanced Study, Hangzhou, 310000, China (e-mail: yxdong@tias.ac.cn).

Zhenyu Zhang, School of Intelligence Science and Technology, Nanjing University, Suzhou, 215000, China (e-mail: zhangjesse@foxmail.com).

Corresponding authors: Jian Yang and Zhenyu Zhang.}
\thanks{}}

\markboth{Journal of \LaTeX\ Class Files,~Vol.~14, No.~8, August~2021}%
{Shell \MakeLowercase{\textit{et al.}}: A Sample Article Using IEEEtran.cls for IEEE Journals}

\maketitle
\begin{figure*}[!t]
    \includegraphics[width=\textwidth]{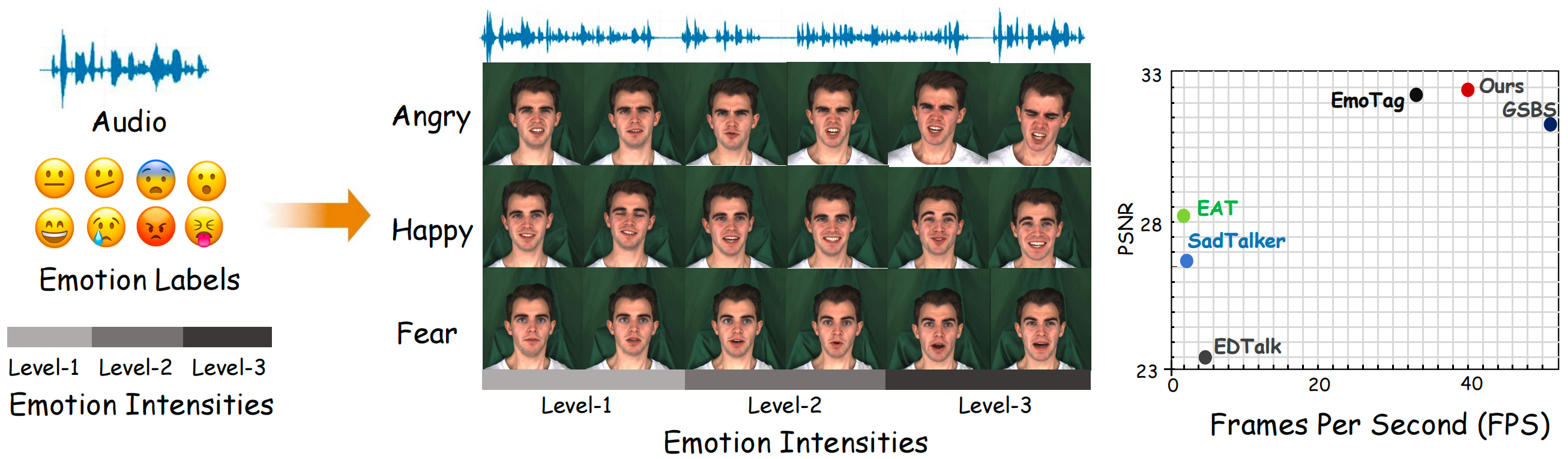}
    \caption{Given the audio, emotion category, and intensity, our method can real-time render high-fidelity, emotion-driven avatars with
 accurate lip-syncing for different emotions and intensities. It outperforms the current state-of-the-art methods in video quality (PSNR) and achieves real-time generation efficiency (FPS).}
    \label{fig:teaser}
\end{figure*}

\begin{abstract}

Audio-driven talking head synthesis has achieved impressive progress in lip synchronization and visual quality, yet generating expressive emotional avatars with controllable intensity remains challenging, especially under real-time constraints. In this paper, we present GaussianEmoTalker, an audio-driven framework for real-time emotional talking head synthesis based on 3D Gaussian Splatting. Instead of directly predicting the final emotional avatar from speech, we formulate emotional animation as a neutral-to-emotional residual deformation problem. GaussianEmoTalker first constructs an identity-specific neutral talking space with GaussianBlendshapes, which provides high-fidelity Gaussian attributes and phoneme-synchronized neutral motion. It then predicts an emotion-conditioned residual deformation by combining mesh displacement cues, audio features, emotion categories, and intensity encodings. To fuse these heterogeneous signals, we introduce a spatial-audio-emotion attention module that estimates the offsets of Gaussian attributes for expressive and temporally stable rendering. Extensive experiments demonstrate that GaussianEmoTalker achieves competitive video quality, accurate lip synchronization, controllable emotional expression, and real-time rendering compared with recent emotional talking head methods. Our project page is available at \url{https://njust-yang.github.io/GaussianEmoTalker.github.io/}
\end{abstract}


\section{Introduction}
\label{sec:intro}
\IEEEPARstart{I}{n} virtual reality and related applications, the significance of animated 3D virtual human heads grows substantially. Among these applications, audio-driven portrait animation plays a key role across various domains, such as human-computer interaction, digital humans, film production, and virtual video conferencing.

With the advent of Generative Adversarial Networks (GANs), earlier methods \cite{SDE,ATFG,ADF,TAI3,TAI4} either directly learn the mapping from audio to video frames or use intermediate representations (e.g., landmarks) to connect audio input with video output. However, these approaches mainly focus on addressing lip-syncing and video quality issues, with few investigating the generation of emotionally expressive videos. Recently, research on emotion-driven virtual head generation has gained significant attention \cite{media2face}. For example, some methods \cite{SDTF,LUM,ECGT} use one-hot encoded emotional labels as the source of emotional data, while EDTalk \cite{EDTalk} decouples lip movements, head poses, and emotions to fit new emotional expressions. However, these methods lack the capability to edit emotional intensity. With the introduction of the multi-intensity emotional dataset MEAD \cite{mead}, several new methods emerge to edit emotional intensity. For instance, EAT \cite{EAT} constructs an emotional deformation network and an emotion-adaptive module to predict shifts in 3D landmarks for generating emotionally expressive faces. However, these methods depend on the accuracy of external detectors and struggle with generalization across different emotional expressions. To address these limitations, EMOdiffhead \cite{zhang2026emodiffhead} leverages the DECA \cite{deca} method to extract facial expression vectors, combines them with audio input, and guides the diffusion model in generating videos with precise lip synchronization and rich emotional expressions. However, due to the inherent nature of diffusion models \cite{diffposetalk}, these methods face challenges related to generation efficiency, which is crucial for real-world applications.


Based on the discussion above, we pose the following question: \textbf{Can we develop an audio-driven emotional avatar editing pipeline that preserves rich emotional expressions while achieving real-time performance?} Emotional talking head editing can be viewed as a specialized form of standard talking head animation, and the most widely used method for achieving real-time talking head generation is 3D Gaussian Splatting \cite{gaussiansplattingrealtime}. Consequently, this approach should also be applicable for emotion-driven avatars. We conceptualize this task as a deformation process from a neutral facial state to an emotional state, with corresponding changes in the 3D Gaussian attributes. Rather than directly regressing the final emotional avatar from audio in a single step, we explicitly decompose the problem into a neutral talking state and an emotion-dependent residual deformation. The intuition is that phoneme-related articulation is shared across emotions and should be anchored first, while emotion mainly acts as a structured residual that modifies brows, cheeks, eyelids, and mouth corners. This decomposition reduces the search space of the second stage and makes the learned Gaussian offsets easier to interpret. The key areas that require attention are: 1) How can we use 3D Gaussians to construct a neutral state space for representing portraits, while accurately capturing the detailed identity features of the portrait? This is a critical foundation for the next stage. 2) How can we effectively control the changes in 3D Gaussian attributes based on emotion categories, intensities, and audio content?


To this end, we propose the GaussianEmoTalker framework, which utilizes 3D Gaussian Splatting to construct both a neutral and an emotional space for talking heads. By learning the mapping between these two spaces, the framework updates the 3D Gaussian attributes to render talking heads that reflect different emotion categories and intensities, aligned with the corresponding audio content. The framework is structured in two stages: Stage 1: We employ GaussianBlendshapes \cite{Gaussianblend} to build the neutral state space for the portrait. This approach effectively captures the high-frequency details of the face, ensuring a solid initialization of the Gaussian attributes. Stage 2: We establish the emotional state space and introduce a spatial-audio-emotion attention module, which predicts Gaussian residuals from mesh displacement, audio, emotion categories, and intensities.

In summary, our contributions are as follows:
\begin{itemize}
\item We propose an audio-driven 3D Gaussian Splatting framework for real-time 3D emotional speech synthesis. 
\item We utilize 3D Gaussians to independently construct state spaces for both neutral and emotional facial expressions, allowing the model to effectively adapt to the mapping between varying audio and emotional inputs.
\item We propose a spatial-audio-emotion attention mechanism that leverages the spatial variations of the mesh, combined with audio and emotional features, to jointly control the deformation of the Gaussian attributes.
\end{itemize}

\section{Related Work}
\subsection{Audio-driven Talking Head Generation}

In recent years, audio-driven talking head generation technology gains significant attention, particularly with advancements in deep learning. This technology analyzes audio signals to generate facial expressions or animations that match the audio content. Early approaches to audio-driven talking head generation \cite{yousaid,TFGC,ASA,TFGD,VOCA} use encoder-decoder architectures to successfully synthesize talking head videos. To improve lip synchronization, several studies \cite{LSE,TFGC,TFGD} extract disentangled appearance and semantic features from speech or train specialized evaluation networks to optimize lip movement synchronization. With the introduction of generative models, researchers begin generating talking heads directly from extracted features or intermediate representations, such as facial landmarks \cite{HCM,HI,faceformer,codetalker}. Most methods \cite{VDUB,MLS,NVP} synthesize only the mouth region in the video and blend it into the full frame without altering other areas. In contrast, some approaches \cite{SObama} introduce re-timing strategies to find the optimal frame for matching the predicted mouth shape, or use facial landmarks as intermediate representations \cite{ADE,LSP} in image-to-image translation networks to generate full-head animations. However, these methods struggle to generate natural and consistent head poses due to the lack of explicit 3D structural information. GaussianSpeech \cite{Gaussianspeech} further employs 3D Gaussian Splatting to construct high-fidelity, lightweight 3D head avatars, supporting multi-view consistency and real-time rendering, but due to the limitations of tracked FLAME, it cannot capture exaggerated facial expressions and also lacks the ability to render emotion-driven portraits. More recently, EmoTaG \cite{EmoTaG} extends Gaussian-based talking heads to emotion-aware few-shot personalization by combining FLAME-guided Gaussian animation with a gated residual motion network. Unlike our method, however, EmoTaG follows a pretrain-and-adapt few-shot setting and relies on auxiliary pose-expression cues during inference, whereas our method focuses on identity-specific real-time emotional synthesis under the person-specific monocular setting.

\subsection{NeRF-based Talking Head Generation}
In recent years, methods\cite{TAI1,TAI2} based on Neural Radiance Fields (NeRF)\cite{NERF} have attracted considerable attention due to their ability to learn and render the shape and appearance of objects from various viewpoints using neural networks. Notably, ADNeRF \cite{ADNERF} is the first to apply NeRF for synthesizing audio-conditioned talking head scenes. However, the head and torso are modeled separately using two NeRF models, leading to inconsistent results. To resolve this issue, SSP-NeRF \cite{SAI} integrates a facial parsing branch with a torso deformation module, improving rendering efficiency and modeling non-rigid torso deformations within a unified neural radiance field. Additionally, DFA-NeRF \cite{DFA} decouples head pose, eye blink, and lip motion, enabling more personalized animation synthesis. To speed up training, DFRF \cite{DFRF} develops a base model for lip motion that can be quickly fine-tuned to different identities, allowing for few-shot talking head synthesis. ER-NeRF \cite{ER-NERF} optimizes spatial contributions through a tri-plane hash representation. In contrast to these approaches, we leverage 3D Gaussian Splatting \cite{gaussiansplattingrealtime} to model dynamic facial motion, capturing transitions from neutral to emotional states and achieving superior visual quality with real-time rendering.

\subsection{Emotional Talking Head Generation} Emotional expression plays a crucial role in generating talking head animations, as it significantly enhances the realism of the animation. However, many earlier studies focus primarily on the issue of audio-to-lip synchronization, while neglecting the generation of facial expressions. Eskimez et al. \cite{SDTF} and Sinha et al. \cite{ECGT} use one-hot encoded emotion labels to generate emotionally expressive talking faces, but this approach cannot adjust the intensity of the emotion. To address this limitation, Wang et al. \cite{mead} introduce the MEAD dataset, which contains talking video clips with various emotions and intensities. Based on this dataset, EVP \cite{ADE} proposes a method that decouples content and emotional information from the audio signal, enabling implicit control of emotional intensity in the synthesized video. However, this approach struggles with unseen subjects and audio. Additionally, some studies rely on emotion labels \cite{SDTF,ESD,EAT} or extract facial expressions from additional emotional videos \cite{eamm,ethg,CME,emmn,SSM} for emotion retrieval and synthesis. Recent 3D face reconstruction frameworks, notably SMIRK \cite{SMIRK} , have improved upon methods like DECA \cite{deca} and EMOCA \cite{emoca} by introducing a spatial modeling approach that enhances generalization for diverse and rarely observed expressions. Furthermore, recent 2D generative models like VASA-1 \cite{vasa} have achieved remarkable realism in facial dynamics. However, these 2D methods often lack explicit 3D structural constraints, making it challenging to maintain consistent geometry under large pose variations. With the advent of diffusion models, the EMOdiffhead \cite{zhang2026emodiffhead} method uses DECA to extract facial expression vectors, combines them with audio input, and guides the diffusion model to generate videos with precise lip synchronization and rich emotional expressions. EmoTaG \cite{EmoTaG} further shows that emotional talking heads can be produced with Gaussian splatting under a few-shot personalization setting by predicting FLAME-space motion and emotion-aware Gaussian refinements. However, its pretrain-and-adapt pipeline, auxiliary upper-face cues, and few-shot inference protocol differ substantially from our identity-specific real-time setting. In contrast, our proposed method, GaussianEmoTalker, generates emotionally rich talking head animations in real-time. It not only improves generation efficiency but also ensures precise and diverse emotional expression.

\begin{figure*}[ht!]
    \centering
    \includegraphics[width=\textwidth]{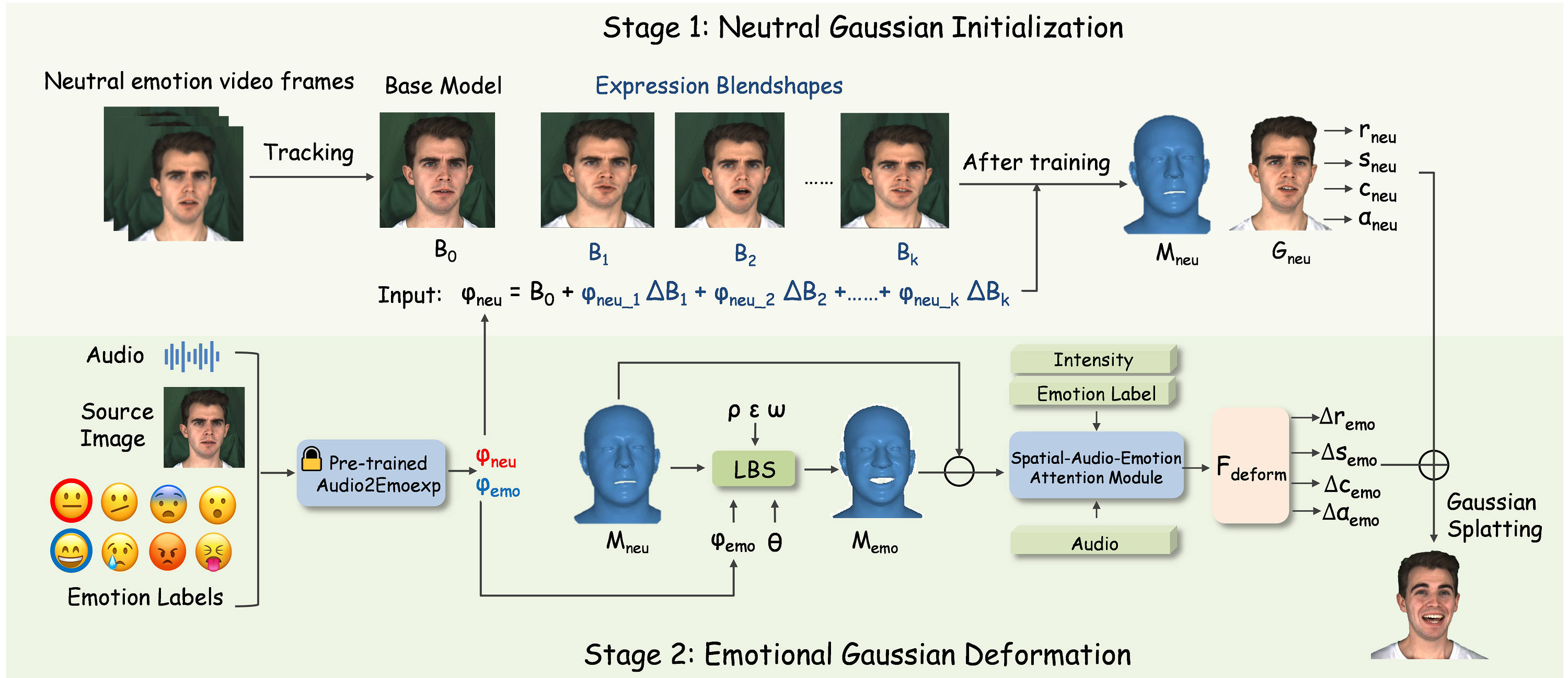}
    \caption{
    \textbf{Overview}. GaussianEmoTalker uses the expression basis of GaussianBlendshapes to construct the neutral state space for the talking head (Stage 1). Through a pre-trained audio-to-expression model, it obtains the neutral expression coefficients and emotional expression coefficients. The former is initialized with the corresponding mesh and Gaussian attributes obtained in Stage 1. The latter is derived from the LBS (Linear Blend Skinning) in the emotion state space constructed by 3D Gaussian, resulting in a deformed mesh. The difference between the two meshes, together with the audio, emotion, and intensity labels, is processed by cross-attention to obtain the final Gaussian attribute offset for the specified emotion, generating the final emotional talking head (Stage 2).}
    \label{fig:fig2}
\end{figure*}

\section{Preliminary}
\subsection{3D Gaussian Splatting} 
Gaussian Splatting uses point clouds to model a scene through a set of 3D Gaussians. Each Gaussian is represented as an ellipse, characterized by its color, opacity, and centered position $x$, which is defined by the covariance matrix $\Sigma$ as follows: $ G(x)=e^{-\frac{1}{2}x^{T}\Sigma^{-1}x}$. For rendering, the 3D Gaussians are projected onto 2D planes using a splatting method. This projection involves a new covariance matrix, $\Sigma^{\prime}$, in camera coordinates, which is defined by: $\Sigma' = JW \Sigma W^TJ^T$, where $W$ is the specified view transformation matrix, and $J$ is the Jacobian matrix corresponding to the affine approximation of the projective transformation. The entire training process is essentially the optimization of the covariance matrix.

\subsection{3D Gaussian BlendShapes} 
The Gaussian blendshape representation consists of a neutral base model ${B_0}$ and a set of expression blendshapes $\{B_1, B_2, \dots, B_K\}$. These bases are derived through principal component analysis based on the FLAME \cite{flame} model. Each model is represented as a set of 3D Gaussian distributions, characterized by several fundamental properties, including position $x$, opacity $\alpha$, rotation $r$, scale $s$. The deviation between ${B_k}$ and ${B_0}$ is defined as the difference in their Gaussian properties, expressed as $\Delta B_k = B_k - B_0$. The head avatar model for a new expression $\psi_k$ can be computed as follows:
\begin{align}
B_\psi = B_0 + \sum_{k=1}^{K} \psi_k \Delta B_k.
\end{align}
Gaussian blendshape are fitted from video, but this can only support more common emotional videos, as it is difficult to fit for strong emotions.

\section{Method}
In this section, we present GaussianEmoTalker, a framework designed for the real-time generation of high-quality virtual avatars driven by both audio and emotion. An overview of the framework is provided in Fig. \ref{fig:fig2}. First, given a video clip of a character speaking in its neutral state, we apply the GaussianBlendshapes \cite{Gaussianblend} to construct the character’s neutral state space. Next, a pre-trained audio-to-expression model is used to generate the corresponding facial expression from the input audio. This expression is then mapped into the neutral state space to produce the initial mesh and Gaussian properties. We then create an emotional state space using 3D Gaussians. Through a cross-attention module that integrates spatial, audio, and emotional information, we calculate the offset of the Gaussian properties relative to the neutral state space, allowing us to render the final emotional avatar. The design is intentionally coarse-to-fine: a neutral branch first captures speech articulation, and a second branch only needs to model the emotion-dependent residual deformation. Finally, we provide a detailed description of the training objectives for this framework.

\subsection{Neutral Gaussian Initialization (NGI)}
As mentioned above, to construct the character's neutral state space, we use GaussianBlendshapes \cite{Gaussianblend} to fit the expressions of each frame in the input neutral video using different expression bases. Then, we use a pre-trained audio-to-expression model to generate new neutral expressions, which are fed into neutral state space to obtain the initial neutral mesh and neutral Gaussian attributes for the audio.

\textbf{Data Preparation.} Following the method of \cite{Gaussianblend}, we utilize the face tracker from \cite{tracker} to compute the FLAME mesh for the neutral expression, along with the meshes for 50 basis expressions derived from input neutral video clips.

\textbf{Gaussian Initialization.} Upon finishing the initial phase of data preparation, we move on to setting up the neutral model $B_0$, the expression blendshapes $\{B_k\}$, and the mouth interior Gaussians $B_m$, as detailed in GaussianBlendShapes \cite{Gaussianblend}. For $B_0$, we place a collection of points on the neutral FLAME mesh $M_0$ through Poisson disk sampling. These points act as the starting positions for the Gaussians. The remaining attributes of the Gaussians are initialized according to the 3DGS methodology. For every Gaussian, we also determine its nearest triangle on $M_0$ and calculate its LBS blend weights by linearly interpolating the blend weights at the triangle's vertices. To set up the mouth interior Gaussians $B_m$, we employ two predefined billboards to depict the upper and lower teeth. These billboards are subsequently sampled into Gaussians using Poisson disk sampling. The Gaussians representing the upper teeth are firmly attached to the rear of the head, whereas those for the lower teeth are connected to the vertex with the highest skinning weight for the jaw joint. Lastly, to initialize the expression blendshapes $B_k$, we apply the deformation gradients from $M_0$ to the expression FLAME mesh $M_k$ to transform each Gaussian of $B_0$.


\textbf{Audio-to-Expression Transformer (A2ET).} Building on EAT \cite{EAT}, we reuse its audio encoder and emotion-conditioning backbone, but replace the original landmark decoder with lightweight regression heads for FLAME coefficients. Specifically, the adapted A2ET predicts a FLAME expression vector $\psi$ and a local pose vector $\theta=(\theta^{jaw},\theta^{neck})$ containing jaw and neck rotations only. The source image $I_s$ is a fixed canonical neutral frame of the target identity and is used in both training and inference to preserve identity-specific mouth geometry when mapping audio to FLAME coefficients. The model also accommodates emotion label inputs, as described by the following equations:
\begin{align}
\mathcal{F}_{A2ET}(A,I_s,neu)\to \psi_{neu},\theta_{neu},
\end{align}
\begin{align}
\mathcal{F}_{A2ET}(A,I_s,emo)\to \psi_{emo},\theta_{emo},
\end{align}
where $A$ refers to audio, $neu$ refers to the neutral expression label, and $emo$ represents the other seven emotion labels. Since MEAD is captured with an almost fixed viewpoint, we do not predict global head pose or eye pose. These components are fixed to the canonical values estimated by the tracker for each identity, while $\theta$ only models local jaw and neck motion. Next, the neutral expression $\psi_{neu}$ is fed into the pre-fitted GaussianBlendshapes, generating the initialized neutral mesh $M_{neu}$ and Gaussian attributes $G_{neu}$, as shown below:
\begin{align}
\label{eq4}
\mathcal{F}_{GB}(\psi_{neu})\to M_{neu}, G_{neu}.
\end{align}

\subsection{Emotional Gaussian Deformation (EGD)}
As mentioned earlier, we have established the neutral state space, where variations in emotions, intensities, and audio content are represented as facial offsets relative to the neutral state. Building on this, we have developed an emotional Gaussian deformation space.

\textbf{Neutral Mesh to Emotion Mesh.} Since the other parameters of the Gaussian points are computed from their mean position, and in the first stage, we have obtained the neutral mesh, we depict it as standard points. Referencing IMavatar \cite{imavatar}, we transform the Gaussian points from the neutral space to the emotional deformation space through target emotion expression and pose parameters, based on learned blendshapes and skinning weights. The formula is as follows:

\begin{align}
\mathbf{M}_{emo} = \text{LBS}(\mathbf{M}_{neu} + \mathbf{B_P}(\theta_{emo}; \mathbf{\rho}) + \mathbf{B_E}(\psi_{emo}; \mathbf{\varepsilon}), \\
\mathbf{J}(\psi_{emo}), \theta_{emo}, \mathbf{\omega}), \notag
\end{align}
where $LBS$ and $J$ represent the standard skinning function and joint regressor in FLAME \cite{flame}, while $B_P$ and $B_E$ denote the linear combinations of blendshapes that generate pose and expression, using the animation coefficients $\theta_{emo}$ and $\psi_{emo}$, along with the blendshape bases $\rho$ and $\varepsilon$. The residual motion cue used by Stage 2 is $\Delta M=M_{emo}-M_{neu}$.


\textbf{Disentangling Speech-Related Motion.} When synthesizing a talking head, the corresponding speech audio alone does not capture the full complexity and variety of facial movements. Subtle expressions such as eye blinks, facial wrinkles, and external factors like hair movement and lighting changes are often independent of the speech audio. Therefore, when mapping speech to 3D Gaussian deformations, it is essential to separate non-verbal motions from scene variations. Drawing inspiration from \cite{GaussianTalker}, we introduce auxiliary conditions to capture non-verbal motions, which allows us to disentangle speech-related facial movements from monocular video. Specifically, we use AU45 from the Facial Action Coding System (FACS) to quantify the degree of eye blinking and apply sinusoidal positional encoding to align the input dimensions. We also incorporate the tracked camera parameters $c_n$ of the aligned crop as an auxiliary code to absorb the slight frame-wise viewpoint variation that remains after preprocessing. Finally, we introduce a null token $z_{\emptyset}$, implemented as an all-zero vector with the same dimensionality as the other conditioning channels. For frame $n$, the initial query of the attention module:
\begin{align}
q_n = W_q[f(\Delta M_n)\, \| \, PE(AU45_n)\, \| \, c_n \, \| \, z_{\emptyset}],
\end{align}
where $\|$ denotes concatenation and $W_q$ is a learned linear projection. In the ablation ``w/o Null Vector'', we keep the same projection layer and parameter count, and only mask out $z_{\emptyset}$ so that the architecture remains unchanged. The null token acts as a global anchor that helps the network reserve capacity for motion-independent appearance and long-term dynamics instead of forcing all channels to follow high-frequency audio fluctuations.

\textbf{Emotion and Intensity Text Encoding.} In this study, we use the CLIP (Contrastive Language-Image Pretraining) \cite{clip} model to encode emotion and intensity texts. The CLIP model is built on a dual-tower architecture, consisting of a text encoder and an image encoder. For the purpose of this work, we focus solely on the text encoder to process the emotion and intensity texts.
The emotion text, denoted as $t_e$, includes categories such as ``happy", ``sad", ``fear", ``disgusted", ``contempt", ``angry" and ``surprised". The intensity text, denoted as $t_s$, includes levels such as "level1", "level2", and "level3". Both emotion and intensity texts are encoded using CLIP's TextEncoder, as outlined below:
\begin{align}
f_e=TextEncoder(t_e),f_s=TextEncoder(t_s).
\end{align}

\textbf{Spatial-Audio-Emotion Cross Attention.} Previous approaches for implementing region-aware audio or emotion typically adjust the audio or emotion weights at each 3D spatial point using element-wise multiplication. However, in dynamic talking head scenarios, the variability of audio inputs makes it challenging for a static 3D point to consistently maintain the same audio or emotion weight. This can lead to convergence issues during training, as a fixed 3D coordinate may no longer correspond to the same facial region. To address this problem and enhance the extraction of spatial-audio and emotion features, we propose a spatial-audio-emotion cross-attention module. This module encodes the interpolation between the meshes of the neutral state space and the emotional state space, merging them with subsequent audio and emotion features. This allows the model to capture how the input audio and emotion affect the movement of the 3D Gaussians. The spatial-audio-emotion attention module consists of multiple sets of cross-attention layers $F_{ca}$ and feed-forward layers $F_{fd}$, each connected via skip connections. The module is formulated as follows:
\begin{align}
&z_{n}^{0}=q_n,\\
&z_{n}^{\prime l}={F}_{ca}(z_{n}^{l-1},a_{n},f_e,f_s)+z_{n}^{l-1},\quad l=1...L,\\
&z_{n}^{l}=F_{fd}(z_{n}^{\prime l})+z_{n}^{\prime l},\quad l=1...L.
\end{align}
 By computing the cross-attention between the spatial query $q_n$, audio features $a_{n}$, emotion features $f_e$, and intensity features $f_s$ of the $n^{th}$ image frame, the output features successfully integrate these signals with the rich facial details captured by the Gaussian model, preparing for subsequent Gaussian deformation in the emotion space.

\textbf{Gaussian Deformation.} Previous methods rely on a conditional NeRF \cite{NERF} representation, where the 3D coordinates of sampling points along each ray remain fixed, and only color and density are conditioned on the input audio. In contrast, using an explicit representation of 3D Gaussians enables us to manipulate not only the appearance but also the spatial positions and shapes of each Gaussian primitive. This approach leads to rendered images with richer details and improved quality. We have developed a deformation network that utilizes cross-attention across spatial, audio, and emotional inputs to compute deformations of Gaussian attributes, such as color, rotation, and scaling. To predict the offsets for each Gaussian attribute, we use a set of MLP regressors, $\mathcal{F}_d$, as detailed below:
\begin{align}
(\Delta r_{emo},\Delta s_{emo},\Delta c_{emo}, \Delta \alpha_{emo})=\mathcal{F}_d(z_{n}^{l}).
\end{align}
Therefore, the Gaussian attributes in the final emotional space are: $r_{emo} = r_{neu} + \Delta r_{emo}$, $s_{emo} = s_{neu} + \Delta s_{emo}$, $c_{emo} = c_{neu} + \Delta c_{emo}$, $o_{emo} = o_{neu} + \Delta o_{emo}$ representing rotation, scale, color, and opacity, respectively. $r_{neu}$, $s_{neu}$, $c_{neu}$, $o_{neu}$ are derived from $G_{neu}$ in equation (\ref{eq4}).

\noindent\textbf{Gaussian Splatting Rendering.} In the previous section, we obtain all the Gaussian parameters in the emotional deformation space, which allows us to use these parameters for rendering. We overlap $N$ points in order of depth onto a single pixel, forming the blended color of the pixel, as expressed below:
\begin{align}
C_{\mathrm{pix}}=\sum_{i\in{\cal S}}c_{emo}{}^{i}\Pi(f_{emo}{}^{i})\prod_{j=1}^{i-1}(1-\Pi(f_{emo}{}^{j})), 
\end{align}
where $\Pi(f_{emo}{}^{i})$ is the influence of each Gaussian point on the  pixel, and $\prod_{j=1}^{i-1}(1-\Pi(f_{emo}{}^{j}))$ demonstrates the transmittance term.

\subsection{Training Objectives}
The two stages of our framework output a rendered image, so both stages use the same loss function. Similar to prior work \cite{imavatar}, the RGB loss serves as a supervisory signal, ensuring alignment with the color of each pixel.
\begin{align}
\mathcal{L}_{\mathrm{RGB}}(\mathrm{I_{GS}})=\|\mathrm{I_{GS}}-\mathrm{I}_{\mathrm{GT}}\|,
\end{align}
where $I_{GS}$ and $I_{GT}$ are the predicted image and ground truth image. Additionally, we use the VGG \cite{vgg} feature loss because Gaussian splatting offers high rendering efficiency and can render all images at each step:
\begin{align}
\mathcal{L}_{\mathrm{vgg}}(\mathrm{I_{GS}})=\|\mathrm F_{\mathrm{vgg}}(\mathrm I_{GS})-\mathrm F_{\mathrm{vgg}}(\mathrm I_{\mathrm G\mathrm T})\|,
\end{align}
where $F_{vgg}$ denotes the feature outputs from the first four layers of a pre-trained VGG network. We also use the flame loss:
\begin{align}
\mathcal{L}_{\mathrm{flame}}&=\frac{1}{N}\sum_{i=1}^{N}(\lambda_{e}\|\mathcal{E}_{i}-\widehat{\mathcal{E}}_{i}\|_{2}+\lambda_{p}\|\mathcal{P}_{i}-\widehat{\mathcal{P}}_{i}\|_{2}\\&+\lambda_{w}\|\mathcal{W}_{i}-\widehat{\mathcal{W}}_{i}\|_{2}),
\end{align}
the FLAME loss leverages prior knowledge about expression and pose deformations from FLAME, by supervising the deformation network with the corresponding values of the nearest FLAME vertices, where $\mathcal E_i$, $\mathcal P_i$, and $\mathcal W_i$ denote the predicted values of the deformation network, $\widehat{\mathcal{E}}$, $\widehat{\mathcal{P}}$, and $\widehat{\mathcal{W}}$ represent pseudo ground truth values, which are determined based on the nearest FLAME vertex. The final loss is expressed as the combination of a D-SSIM term:
\begin{align}
\mathcal{L}&=\lambda_{\mathrm{rgb}}\mathcal{L}_{\mathrm{RGB}}+\lambda_{\mathrm{flame}}\mathcal{L}_{\mathrm{flame}}+\lambda_{\mathrm{vgg}}\mathcal{L}_{\mathrm{vgg}}\\&+\lambda_{\mathrm{D-SSIM}}\mathcal{L}_{\mathrm{D-SSIM}}.
\end{align}

\section{Experiment}
\subsection{Setup}
\textbf{Datasets.} We train and evaluate our GaussianEmoTalker method using the MEAD dataset \cite{mead}, a high-quality emotional talking-head video dataset. It includes some actors and covers 8 emotion categories, each with three intensity levels. Each actor video is cropped and resized to 512×512. 
For each identity, we partitioned the collection of independent video clips into training and testing sets at a ratio of 9:1. This clip-level split ensures that the testing set consists of entirely unseen sequences, maintaining the integrity of the evaluation across all 8 emotions and 3 intensity levels. Therefore, each training session is based on monocular data.

\textbf{Evaluation Criteria.} Our comparisons are structured into two
 distinct settings: \textbf{self-driven} and \textbf{cross-driven}. In the self-driven setting, we use the following metrics to evaluate the performance of the proposed method in video quality, and lip synchronization. Video quality includes: FID (Fréchet Inception Distance) \cite{FID}, FVD (Frechet Video Distance) \cite{FVD}, PSNR (Peak Signal-to-Noise Ratio) \cite{PSNR}, SSIM (Structural Similarity Index) \cite{SSIM}, LPIPS (Learned Perceptual Image Patch Similarity) \cite{LPIPS}, and CPBD (Complex Plot and Block Distortion)\cite{CPBD}. In addition, we extract facial landmarks from the aligned result sequences and the ground truth sequences, involving two metrics: Landmark Distance (LD) and Landmark Velocity Difference (LVD). These metrics are used to evaluate facial motions. Specifically, LD represents the average Euclidean distance between the generated landmarks and the recorded landmarks. Velocity refers to the change in landmark positions between consecutive frames, so LVD represents the average velocity difference of landmark motions between the two sequences. We apply LD and LVD to the mouth and facial regions. In the cross-driven setting, to evaluate lip synchronization, all methods are guided by entirely distinct audio tracks. Since ground truth images are unavailable, we measure lip sync accuracy through Landmark Distance (LMD) and SyncNet confidence scores (Sync) \cite{SyncNet}. Additionally, the accuracy of facial movements is quantified using Action Unit Error (AUE).

\textbf{Implementation Details.} Our framework is implemented in PyTorch and runs on an RTX 3090 GPU. It consists of two stages. In the first stage, we adhere strictly to the training procedure of GaussianBlendshapes \cite{Gaussianblend}. In the second stage, we continue training the network for 5000 iterations. All modules are trained with an initial learning rate of 0.0001, which gradually decays to 0.00001. For all experiments, we set $\lambda_\mathrm{RGB}=1$, $\lambda_\mathrm{D-SSIM}=0.25$, $\lambda_\mathrm{flame}=1$, and 
$\lambda_\mathrm{vgg}=0.1$. For the flame loss, we use $\lambda_\mathrm{e}=1000$, $\lambda_\mathrm{p}=1000$, and $\lambda_\mathrm{w}=1$. The training process is optimized using the Adam \cite{adam} optimizer.

\textbf{Baselines.} We compare GaussianEmoTalker with SadTalker \cite{sadtalker}, EAT \cite{EAT}, EDTalk \cite{EDTalk}, GSBS \cite{Gaussianblend}, and the recent emotion-aware Gaussian baseline EmoTaG \cite{EmoTaG}. Among these, GSBS is the baseline we have improved upon. 
\begin{figure}[h]
  \centering
    \includegraphics[width=\linewidth]
    {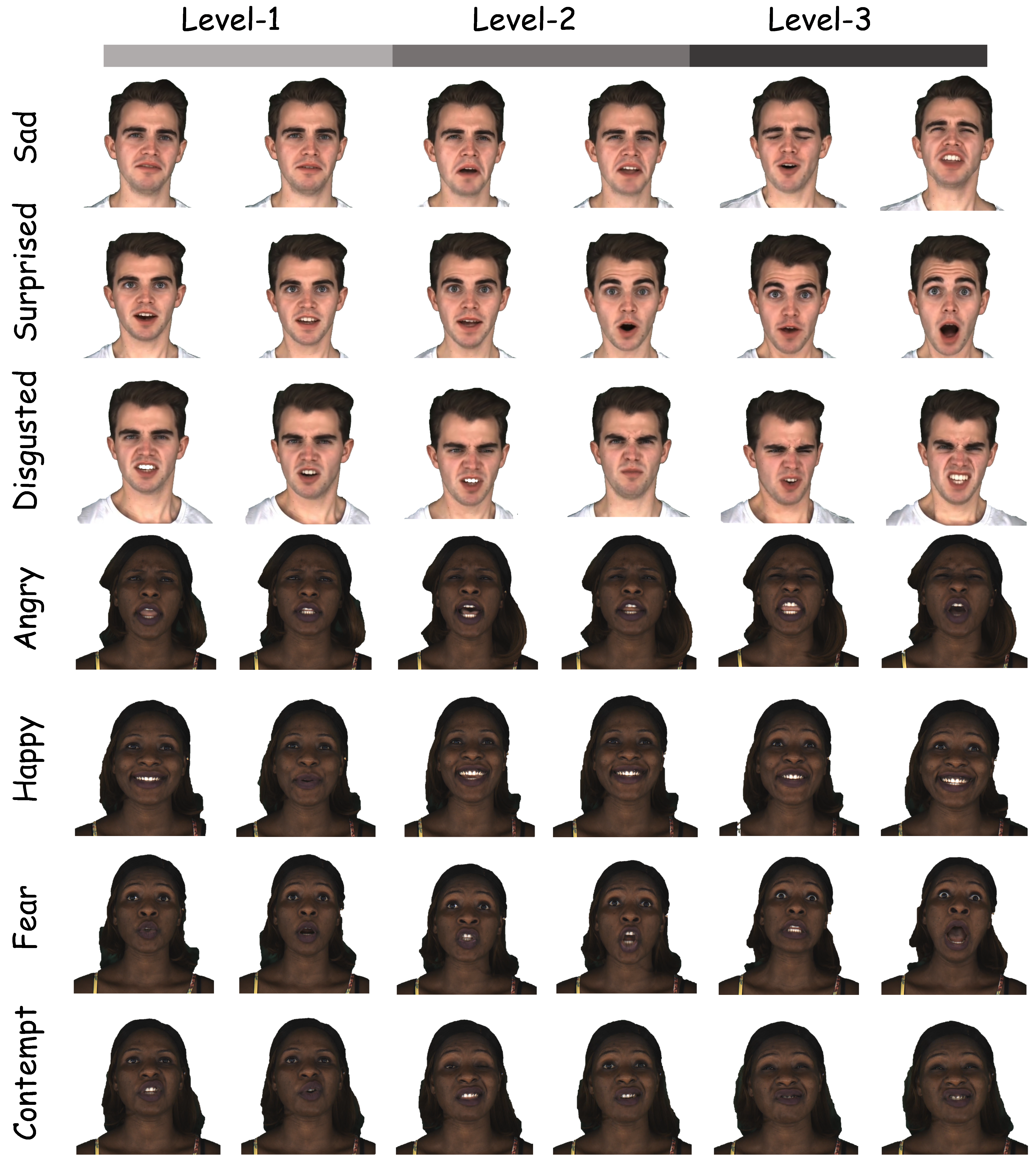}
    \caption{The results of different facial expression types and
intensities generated by our method.}
    \label{fig:result1}
\end{figure}

\begin{figure}[h]
  \centering
    \includegraphics[width=\linewidth]
    {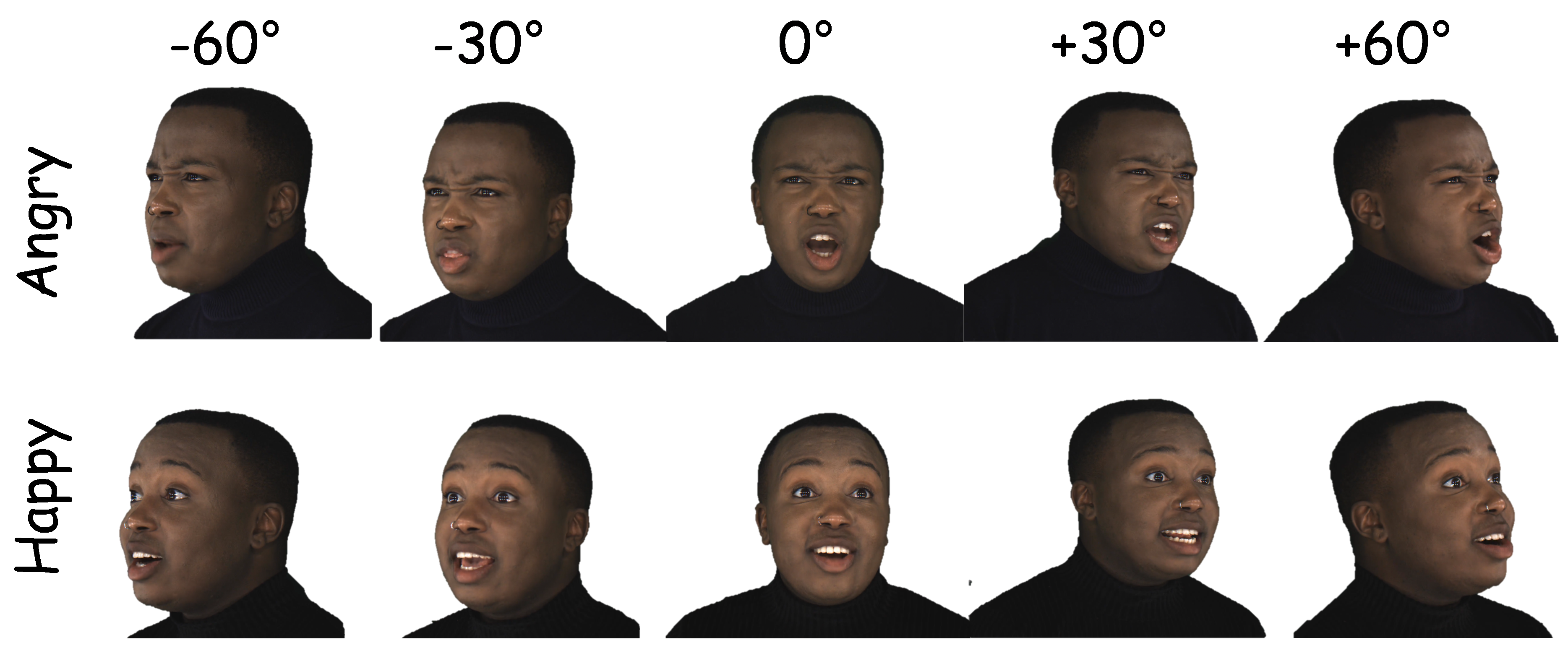}
    \caption{Multi-view synthesis and emotional control.}
    \label{fig:result2}
\end{figure}

\begin{figure}[h]
  \centering
    \includegraphics[width=\linewidth]
    {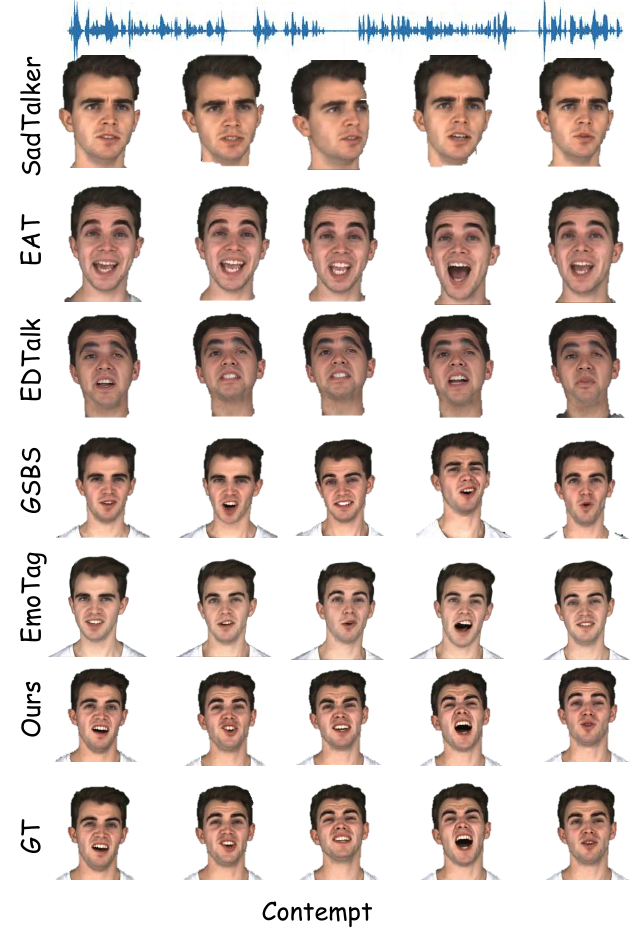}
    \caption{Qualitative comparison in the self-driven setting. All methods are evaluated on the same fixed-view target sequence. The comparison focuses on local lip motion, emotional expression, identity details, and rendering quality.}
    \label{fig:result3}
\end{figure}

\begin{figure}[h]
  \centering
    \includegraphics[width=\linewidth]
    {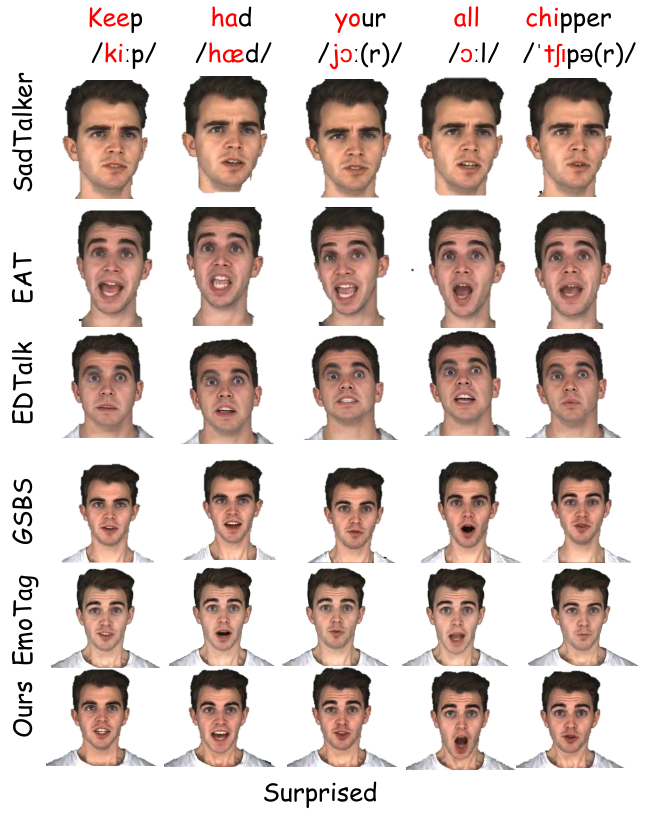}
    \caption{Qualitative Comparison in the cross-driven setting.  The sequence depicts the lip shape conforming to specific phonemes in the spoken words ``keep", ``had", ``your", ``all", ``chipper". Our method outperforms previous works in both lip synchronization and rendering quality.}
    \label{fig:result4}
\end{figure}

\subsection{Qualitative Evaluation}

In the \textbf{self-driven} setting, we qualitatively analyze the emotion editing capabilities of the proposed method, GaussianEmoTalker, and present editing results for different emotion types and intensities shown in Fig. \ref{fig:result1}. The renderings of our results at different views with different emotions are shown in Fig. \ref{fig:result2}. Additionally, we compare our method with several state-of-the-art approaches, including EAT \cite{EAT}, EDTalk \cite{EDTalk}, GaussianBlendShapes (GSBS) \cite{Gaussianblend} and EmoTaG \cite{EmoTaG}, with GSBS representing the first stage of our approach. As shown in Fig. \ref{fig:result3}, our method produces more realistic emotional portraits that better align with the ground truth, especially around lip closure, cheek lifting, and the high-frequency details near the eyes. All qualitative results are composited back into the aligned MEAD crop using the original static background plate of the corresponding target clip, the background is therefore not synthesized by our network and is kept identical across compared methods for readability.


In the \textbf{cross-driven} setting, we also compare our method with other advanced methods. For lip movements corresponding to different words, our results achieve better lip-sync accuracy, as shown in Fig. \ref{fig:result4}. Compared with the baselines, our method preserves tighter lip closure for plosive phonemes and more stable mouth corners across frames. This is attributed to our Spatial-Audio-Emotion cross-attention mechanism, which effectively links audio features with the Gaussian structure while keeping the emotional deformation anchored to the neutral talking state.


\subsection{Quantitative Evaluation}
In the \textbf{self-driven} setting, we evaluate video quality and lip synchronization for each method using specific metrics. As demonstrated in Tab. \ref{tab:1}, our method outperforms all others across these metrics. Notably, since our approach is based on Gaussian splatting, it achieves high generation efficiency, enabling real-time performance, as illustrated in Fig. \ref{fig:teaser}.

In the \textbf{cross-driven} setting, we evaluate lip synchronization for each method using specific metrics. As demonstrated in Tab. \ref{tab:2},
We utilizd four segments of audio unrelated to individuals and conducte comparisons with other methods under three different metrics $Sync$, $LMD$, $AUE$. These clips do not exist in the test set but were randomly selected from audio clips of other characters. We tested all emotional labels and their intensities, and then averaged the results. It can be observed that our approach has achieved state-of-the-art performance across the board.

In addition, to enhance the generalization ability of our method, we jointly train the model using all identity subjects. Although the performance slightly decreases compared to training on individual identities, it still achieves better results than other methods.

\textbf{User Study.} we conduct a user study to more precisely evaluate the visual quality of emotion-driven talking head videos. The study involves 20 participants, including master’s and PhD students with a background in computer vision, who assess the rendered results within the head reconstruction framework. To facilitate accurate evaluation, we compile all the generated videos into a single high-resolution video, enabling participants to observe all movements simultaneously. To ensure fairness, each video is assigned a number, rather than being identified by its corresponding method. Our evaluation focuses on intensity level 3 (the most distinct expressions) across all 8 emotion categories for 3 representative actors (2 male and 1 female), totaling 24 key test sequences. Participants are asked to evaluate the generated portraits based on four criteria: (1) Emotional accuracy, (2) Audio-visual synchronization, (3) Identity consistency, and (4) Video quality, as shown in Tab. \ref{tab:3}. 

\begin{table*}[t!]\small
    \centering
   \setlength{\tabcolsep}{1mm}
   \caption{
        Quantitative comparison with other state-of-the-art methods in terms of video quality and audio-video synchronization in the self-driven setting. M- represents mouth and F- stands for face region.
        }
    \begin{tabular}{c|c|c|c|c|c|c|c|c|c|c|c|cc}
        \midrule
        \multirow{2}{*}{Method} & \multicolumn{10}{c|}{Video Quality}   &\multicolumn{1}{c|}{Lip Sync} & \multirow{2}{*}{FPS$\uparrow$} \\ 
        &FID$\downarrow$&FVD$\downarrow$&PSNR$\uparrow$&SSIM$\uparrow$&LPIPS$\downarrow$&CPBD$\uparrow$&M-LD$\downarrow$&M-LVD$\downarrow$&F-LD$\downarrow$&M-LVD$\downarrow$&$Sync_{conf}\uparrow$\\
        \midrule
        SadTalker & 49.53 & 344.19 & 26.56 & 0.65 & 0.37 & 0.33 & 2.66 & 2.45& 3.23 & 3.01 &2.56 & 2.2\\
        EAT & 54.22 & 389.74 & 28.39 & 0.72 & 0.36 & 0.41 & 2.90 & 2.66& 3.67 & 3.33& 3.27 & 2.0\\
        EDTalk & 64.76 & 432.90 & 23.43 & 0.61 & 0.43 &  0.29 & 3.11 & 2.88& 3.99 & 3.56 & 2.93 & 4.8\\
        GSBS &  32.33 & 169.32 & 30.78 & 0.75 & 0.27 & 0.41 & 2.57 & 2.10 & 3.18 & 2.56 & 4.34 & \textbf{51.0}\\
        EmoTaG & 27.95 & 136.82 & 32.22 & 0.82 & 0.19 & 0.44 & 2.41 & 1.78 & 3.02 & 1.88 & 5.08 & 33.0\\
        w/o A2ET & 28.54 & 147.75 & 31.32 & 0.77 & 0.25 &0.41 & 2.47 & 1.89& 3.13 & 1.94& 4.80 & --\\
        w/o EGD & 32.40 &165.54 & 30.90 & 0.75 & 0.25 & 0.42 & 2.45 & 1.80& 3.08 & 1.90& 4.46 & --\\
        {w/o Null Vector ($\emptyset$)} &  {28.51} &  {140.32} & {32.11} & {0.81} & {0.23} &  {0.44} & {2.38}&{1.79}& {3.01}& {1.88} &  {4.96}&  {--}\\
    {Ours-Combine} & {27.77} & {134.46} & {32.35} & {0.82} & {0.17} & {0.43} & {2.37} & {1.77} & {3.04} &{1.86} &{5.19} & {--}\\

        \textbf{Ours} & \textbf{26.36} & \textbf{132.34} & \textbf{32.40} & \textbf{0.84} & \textbf{0.18} & \textbf{0.46} & \textbf{2.35} & \textbf{1.75} & \textbf{2.98} & \textbf{1.83} & \textbf{5.23} & 40.0\\

        \bottomrule
    \end{tabular}
    
    \label{tab:1}
    
\end{table*}

\begin{table*}[t!]
    \centering
   \setlength{\tabcolsep}{1mm}
   \caption{
        Quantitative comparison with other state-of-the-art methods in terms of audio-video synchronization in the cross-driven setting.
        }
    \begin{tabular}{c|c|c|c|c|c|c|c|c|c|c|c|c|c}
        \midrule
        \multirow{2}{*}{Method} & \multicolumn{3}{c|}{Audio I} & \multicolumn{3}{c|}{Audio II}  & \multicolumn{3}{c|}{Audio III} & \multicolumn{3}{c}{Audio IV} \\ 
        &Sync$\uparrow$&LMD$\downarrow$&AUE$\downarrow$&Sync$\uparrow$&LMD$\downarrow$&AUE$\downarrow$&Sync$\uparrow$&LMD$\downarrow$&AUE$\downarrow$&Sync$\uparrow$&LMD$\downarrow$&AUE$\downarrow$\\
        \midrule
        SadTalker & 3.813 & 15.234 & 5.125 & 3.727 & 16.044 & 5.315 & 3.933 & 14.852 & 5.075 & 4.013 & 14.574 & 4.935 \\
EAT & 4.121 & 14.565 & 4.847 & 4.034 & 15.274 & 4.961 & 4.246 & 14.427 & 4.786 & 4.365 & 13.883 & 4.672 \\
EDTalk & 3.451 & 16.821 & 5.534 & 3.426 & 17.567 & 5.676 & 3.773 & 16.345 & 5.456 & 3.734 & 15.834 & 5.342 \\
GSBS & 4.762 & 12.817 & 4.345 & 4.459 & 13.521 & 4.456 & 4.648 & 12.347 & 4.234 & 4.149 & 11.931 & 4.123 \\
EmoTaG & 5.176 & 11.598 & 4.033 & 5.104 & 13.471 & 4.102 & 5.388 & 10.732 & 3.891 & 5.317 & 9.711 & 3.776 \\
w/o A2ET & 4.678 & 12.123 & 4.234 & 4.543 & 13.845 & 4.345 & 4.789 & 11.168 & 4.123 & 4.991 & 11.234 & 3.987 \\
w/o EGD & 4.589 & 12.349 & 4.278 & 4.478 & 13.923 & 4.389 & 4.712 & 11.845 & 4.167 & 4.823 & 11.564 & 4.056 \\
 {w/o Null Vector ($\emptyset$)} &  {5.121} &  {11.728} & {4.076} & {5.021} &{13.689} &  {4.152} & {5.302} & {10.985}& {3.976}& {5.187} &  {9.932}& {3.865} \\
{Ours-Combine} & {5.212} &{11.405} & {4.008} & {5.066} & {13.535} & {4.089} & {5.429} &{10.849}& {3.909} & {5.255} & {9.774} & {3.801} \\
        \textbf{Ours} & \textbf{5.235} & \textbf{11.436} & \textbf{3.940} & \textbf{5.164} & \textbf{13.354} & \textbf{4.054} & \textbf{5.431} & \textbf{10.575} & \textbf{3.848} & \textbf{5.453} & \textbf{9.655} & \textbf{3.719}\\
 
        \bottomrule
    \end{tabular}
    
    \label{tab:2}
    
\end{table*}

\begin{table*}[t!]\small
    \centering
    \caption{
        Quantitative comparisons on video quality and audio-video synchronization with other state-of-the-art methods in the cross-driven setting.
    }
   \setlength{\tabcolsep}{1mm}
    \begin{tabular}{c|c|c|c|cc}
        \midrule
        \multirow{2}{*}{Method} & \multicolumn{4}{c}{User Study} \\ 
        &Emotional accuracy$\uparrow$&Lip synchronization$\uparrow$&Identity consistency$\uparrow$&Video quality$\uparrow$\\
        \midrule
        SadTalker & 0\% & 5\% & 10\% & 0\% \\
        EAT & 15\% & 0\% & 10\% & 0\% \\
        EDTalk & 0\% & 5\% & 0\% & 0\% \\
        GSBS &  10\% & 20\% & 25\% & 5\% \\
        EmoTag & 35\% & 10\% &20\% &20\% \\
        \textbf{Ours} & \textbf{40\%} & \textbf{60\%} & \textbf{35\%} & \textbf{75\%}\\
 
        \bottomrule
    \end{tabular}

    \label{tab:3}
    
\end{table*}

\subsection{Ablation Study}
To evaluate the individual components of our method, we perform the following ablation experiments: 1) The impact of the Gaussian initialization derived from the neutral state space constructed in stage 1 on the final results; 2) The impact of the Gaussian deformation module in the emotional state space developed in stage 2 on the final outcomes; 3) The impact of the pre-trained audio-to-expression module on the final results.

\begin{figure}[h]
  \centering
    \includegraphics[width=\linewidth]
    {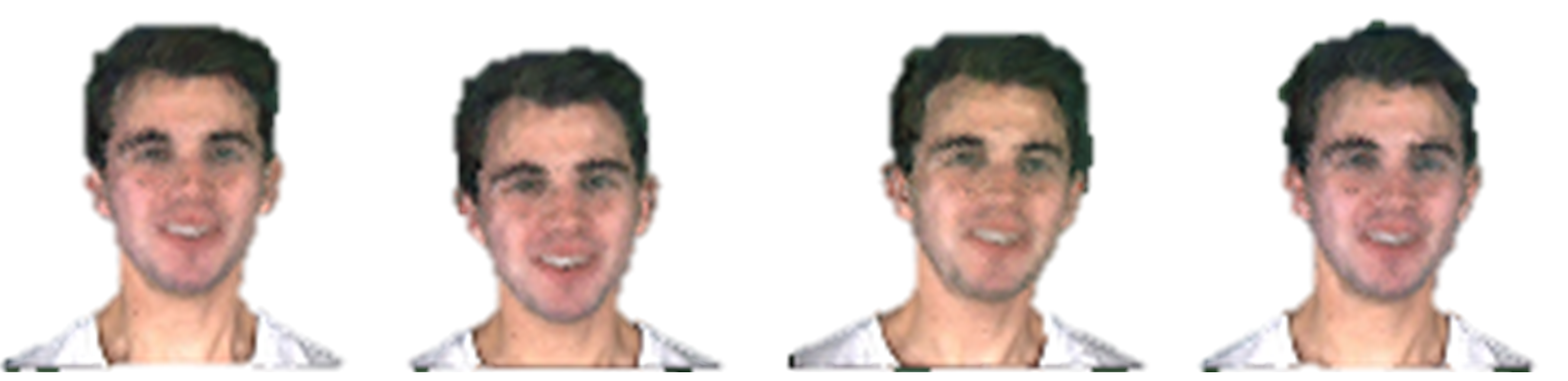}
    \caption{Ablation study results on Gaussian initialization: comparison of Gaussian rendering. The training is stopped at the same epoch when the model without NGI can no longer converge.}
    \label{fig:result6}
\end{figure}

\textbf{Neutral Gaussian Initialization (NGI).} We compare the results with and without Gaussian initialization (i.e., without stage 1), as shown in Fig. \ref{fig:result6}. In our approach, we remove stage 1 and change the output of the stage 2 deformation network to represent the final Gaussian attributes of the emotional space, rather than the offset. As a result, no Gaussian initialization is applied at this stage. The figure shows that training struggles to converge, indicating that directly learning the emotional Gaussian is challenging. This demonstrates the need for a neutral state space to facilitate the transition. The training loss convergence for the aforementioned two settings is illustrated in Fig. \ref{fig:result7}. As can be observed, the training loss exhibits significant oscillations and struggles to converge when NGI is not applied. In addition, to ensure fairness, we compare the training time between with the NGI stage (the two-stage approach of our method) and without the NGI stage (only the second stage), as shown in Tab. \ref{tab:4}. From the table, it can be observed that the NGI stage requires only 0.5 hours and enables the entire model module to converge faster. In contrast, when neutral Gaussian initialization is not performed, the model training requires more time and struggles to converge.
\begin{table}[t!]
    \centering
    \caption{
        Ablation study results on Gaussian initialization: comparison of training time.
    }
   \setlength{\tabcolsep}{1mm}
    \begin{tabular}{c|c|c}
        \midrule
        Method & Stage 1(NGI) & Stage 2 (Emotional Gaussian Deformation)\\ 
        \midrule
        w/o NGI& 0 & $\sim$10h+\\
        with NGI& $\sim$0.5h& $\sim$3h\\
        \bottomrule
    \end{tabular}
    \label{tab:4}
\end{table}

\begin{figure*}[!t]
\centering
\includegraphics[width=0.85\textwidth]{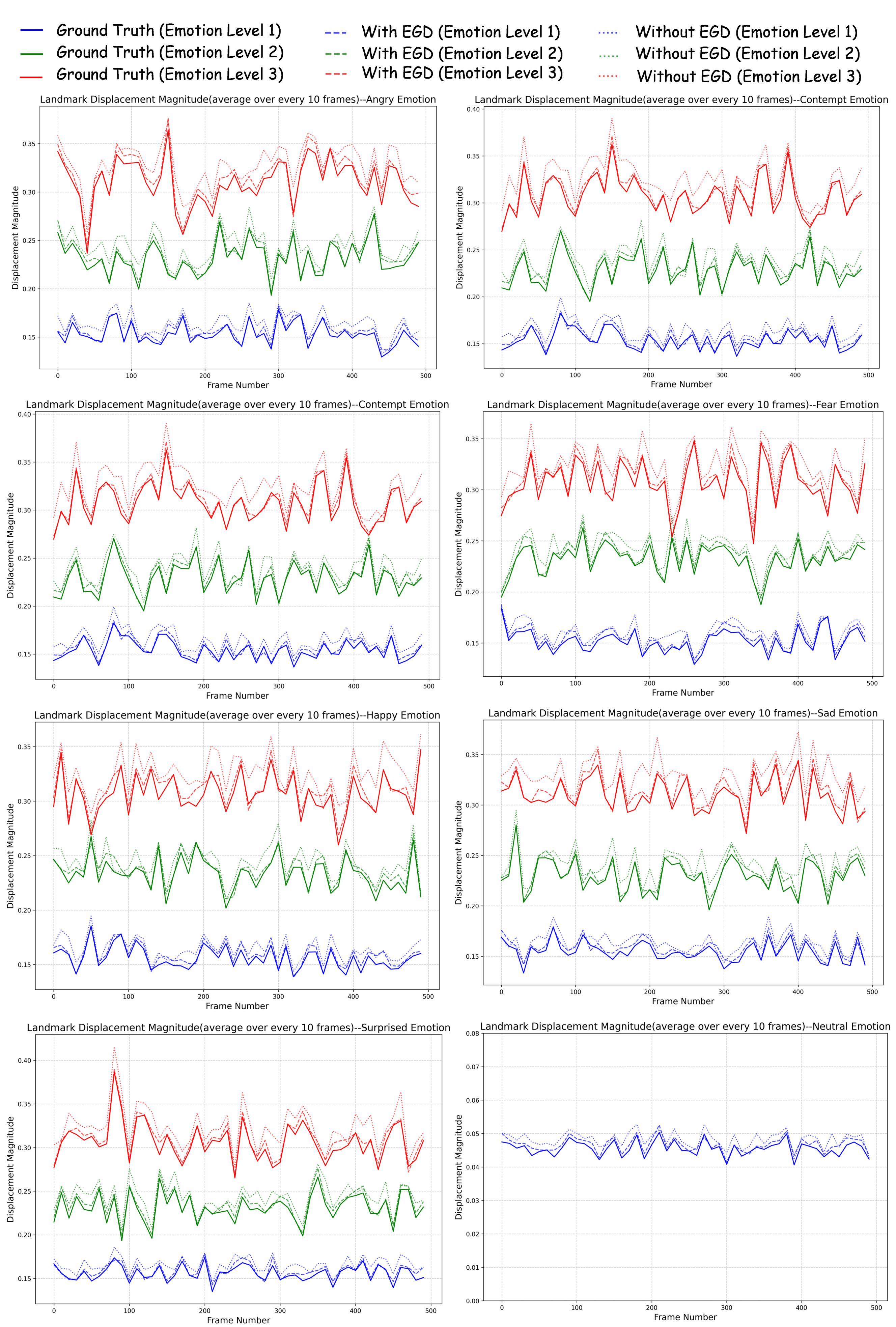}
\caption{Ablation study results on Neutral Gaussian Deformation: comparison of landmark displacement magnitude.}
\label{fig:overall}
\end{figure*}

\begin{figure}[h]
  \centering
    \includegraphics[width=\linewidth]
    {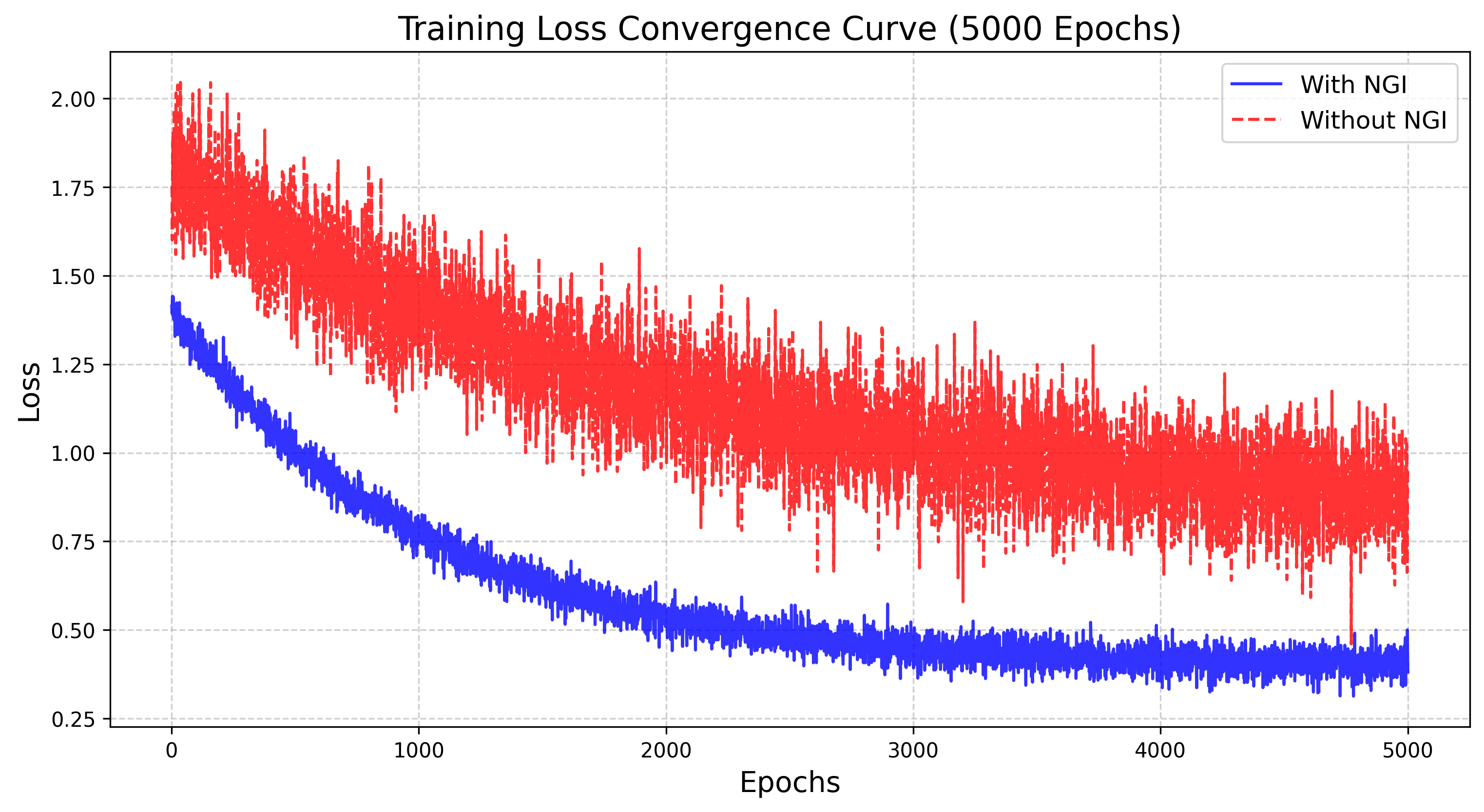}
    \caption{Ablation study results on Gaussian initialization: comparison of training loss.}
    \label{fig:result7}
\end{figure}

\textbf{Emotional Gaussian Deformation (EGD).} We compare the results with and without Gaussian deformation (i.e., without stage 2), as shown in Fig. \ref{fig:result5}. Notably, we did not remove the audio-to-expression module. After fitting the first-stage GaussianBlendshapes to the video (which includes a range of emotions, not just neutral expressions), the emotional expressions generated by the audio-to-expression module are fed back into the first stage for testing. As shown in the figure, without stage 2, the model's ability to express emotions is limited, and it cannot accurately capture exaggerated expressions. This limitation stems from the linear nature of the facial expression base model, which is why we continue to refine and improve upon GaussianBlendshapes. The video quality and audio-lip synchronization metrics are shown in Tab. \ref{tab:1} and Tab. \ref{tab:2}. In addition, to more explicitly compare the fitting capabilities of the two settings for different emotions and their intensity levels, we compare the magnitude of facial landmark displacements with the ground truth. Specifically, we utilize a landmark detector to extract facial key points and then calculate the average displacement magnitude of all key points across every 10 frames (since the changes between consecutive frames are too subtle for effective comparison). As illustrated in Fig. \ref{fig:overall}, when EGD is incorporated, the model demonstrates superior capability in capturing intense emotions (especially at level 3), with displacement magnitudes closer to the ground truth.

\textbf{Audio-to-Expression Transformer (A2ET).} We compare the results with and without the audio-to-expression module, as shown in Fig. \ref{fig:result5}. The module generates expressions by considering both emotional labels and audio content, rather than solely relying on tracking techniques to derive expression coefficients. As a result, it produces more accurate expression coefficients, leading to a more precise implementation of the LBS module and more accurate mesh displacements. As demonstrated in the figure, without the audio-to-expression module, the final emotional expressions are not effectively represented. The results of the quantitative metrics are shown in Tab. \ref{tab:1} and Tab. \ref{tab:2}.

\noindent{\textbf{Null Vector ($\emptyset$).} In this configuration, we remove the globally constant zero-vector input from the spatial-audio-emotion attention module. Our results demonstrate that without this global anchor, the model exhibits significant inter-frame instability and struggles with convergence. This is because the null vector serves as a ``stability anchor", allowing the attention mechanism to effectively decouple motion-independent static attributes (e.g., skin tone, hair color) and long-term dynamic patterns (e.g., breathing rhythms) from high-frequency audio-driven signals. Removing this component leads to noticeable visual jittering, as the network erroneously attempts to map stationary facial regions to fluctuating speech features.

\begin{figure}[h]
  \centering
    \includegraphics[width=\linewidth]
    {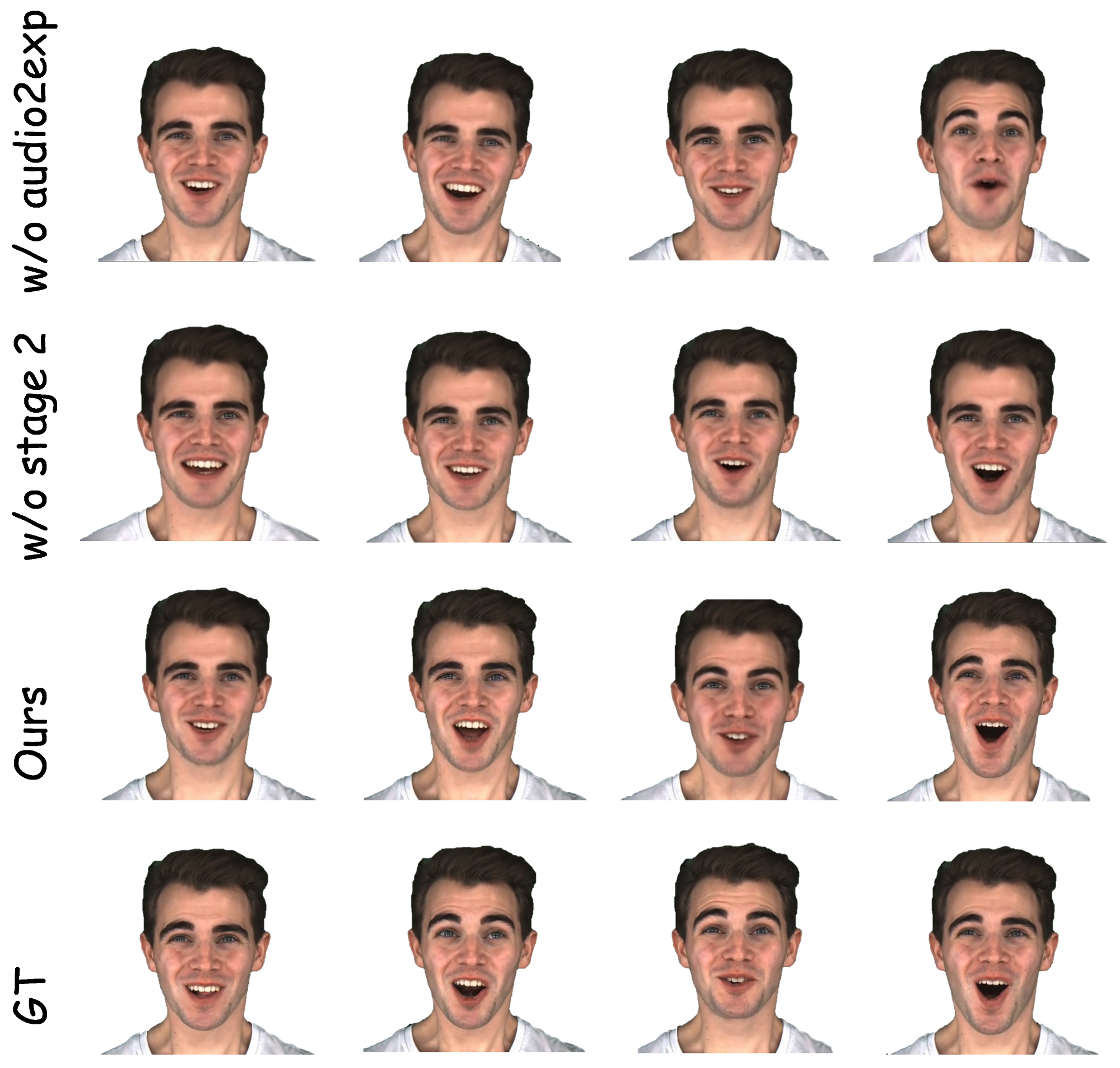}
    \caption{Ablation study results of gaussian deformation and audio-to-expression module.}
    \label{fig:result5}
\end{figure}

\section{Limitation and Future Work}
Although our method is capable of generating expressive talking heads from audio and emotion, there are still two limitations: The first limitation is that our model is person-specific, meaning it can only generate a single identity by training a neural network. This design is specifically tailored for high-fidelity personalized scenarios, such as professional digital avatars and personal digital twins, where preserving unique identity-specific nuances and high-frequency facial details is more critical than cross-identity generalization. Therefore, achieving identity generalization based on Gaussian models would be a promising direction, such as with GAGAvatar \cite{GAGAvatar}. The second limitation is the inability to perform smooth transitions between emotions. In our approach, we can control three discrete levels of emotional intensity, but intermediate intensities, such as 1.5 or 2.5, cannot be achieved. Using linear expressions from facial prior models for interpolation, or applying Gaussian mixture models for interpolation \cite{GMTalker}, are both directions worth exploring.

\section{Discussion}
The proposed framework operates on a strategy of identity-specific monocular training, where each model is trained using video clips captured from a single, consistent viewpoint within the MEAD dataset. By constraining the training to a fixed angle, the framework inherently aligns the generated head pose with the original data, meaning it does not attempt to predict arbitrary global camera trajectories or 3D head movements during inference. While the global relationship between the camera and the subject remains static, anatomical realism is maintained by leveraging the rotational components of the FLAME kinematic chain. Specifically, we utilize the FLAME pose parameters $\theta$ to handle local rotations for the jaw and neck, ensuring that mouth openings and head-neck transitions remain physiologically consistent throughout the animation. The dynamic expressiveness observed in our results, such as the subtle twitching of mouth corners or the characteristic movement of eyebrows, is achieved through the spatial-audio-emotion attention module, which maps audio features and emotion labels to specific Gaussian attribute offsets. Because the model is trained on a person-specific basis, it successfully memorizes the unique micro-expression habits of a particular individual, reproducing these habits deterministically whenever specific audio frequencies and emotional cues are encountered. Furthermore, the introduction of a globally constant zero-vector feature allows the attention mechanism to capture motion-independent static attributes and long-term dynamic patterns, such as breathing rhythms or subtle head inertia, which collectively contribute to the lifelike quality of the output without requiring a stochastic generator.

\section{Conclusion}
GaussianEmoTalker presents a novel framework for the real-time generation and editing of high-quality, audio-driven emotional talking heads. This framework is built on GaussianBlendShapes (GSBS) and initially constructs a neutral state space (Stage 1). It then efficiently creates an emotional Gaussian deformation space using 3D Gaussians, and applies mesh offsets between the neutral and emotional spaces to develop a spatial-audio-emotion cross-attention module. This module learns how different audio content, emotion labels, and emotional intensity affect the Gaussian attribute offsets in the neutral state space (Stage 2). The result is a high-quality emotional talking head rendered in real time.


\bibliography{main}

\bibliographystyle{IEEEtran}

\newpage

\begin{IEEEbiography}[{\includegraphics[width=1in,height=1.25in,clip,keepaspectratio]{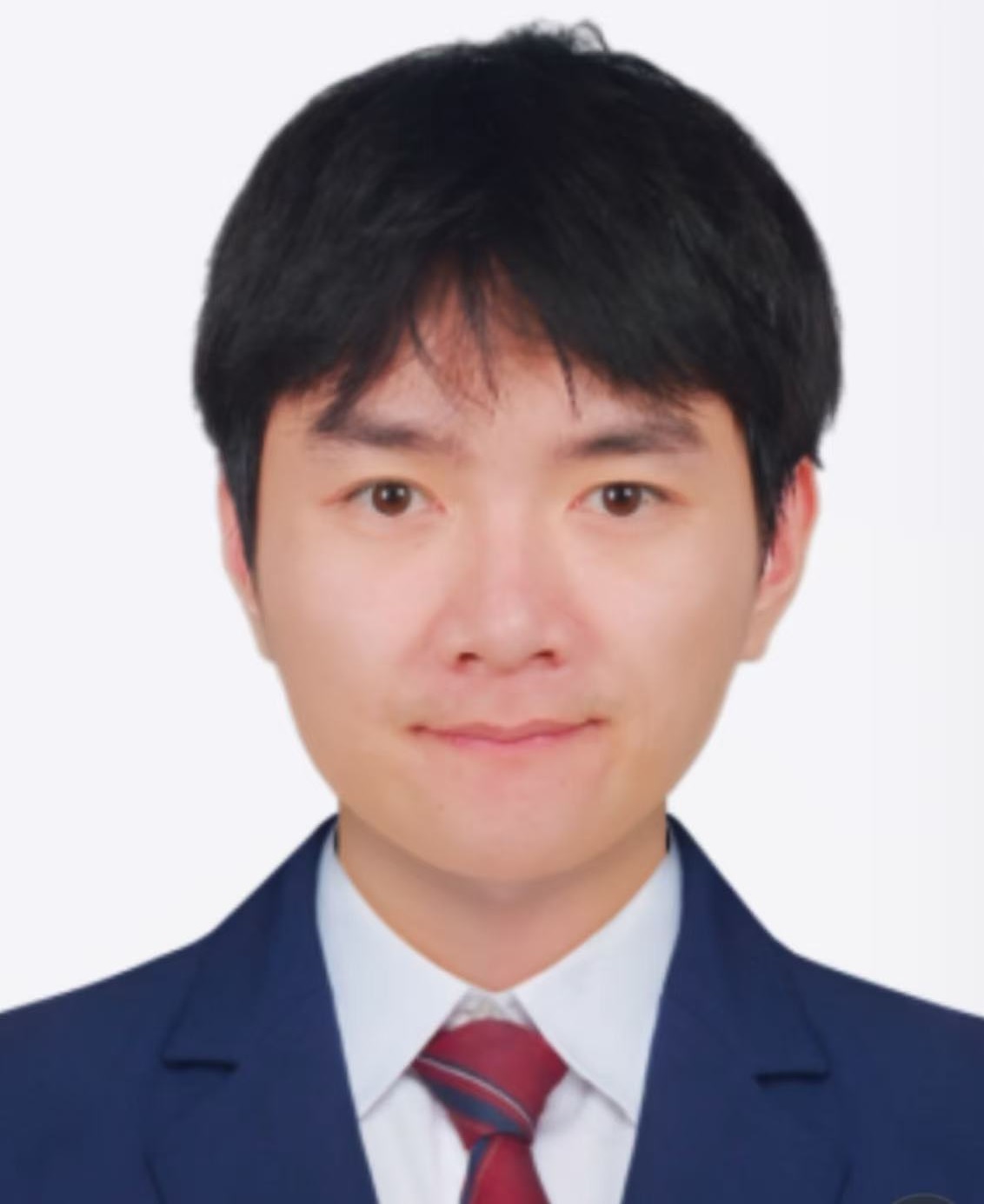}}]{}
\textbf{Haijie Yang} received the M.S. degree in Computer Technology from Hangzhou Dianzi University in 2021. Currently, he is pursuing a Ph.D. in Electronic Information Engineering at Nanjing University of Science and Technology (NJUST) in Nanjing, China. His research interests include 3D reconstruction and pattern recognition, digital human.\end{IEEEbiography}

\begin{IEEEbiography}[{\includegraphics[width=1in,height=1.25in,clip,keepaspectratio]{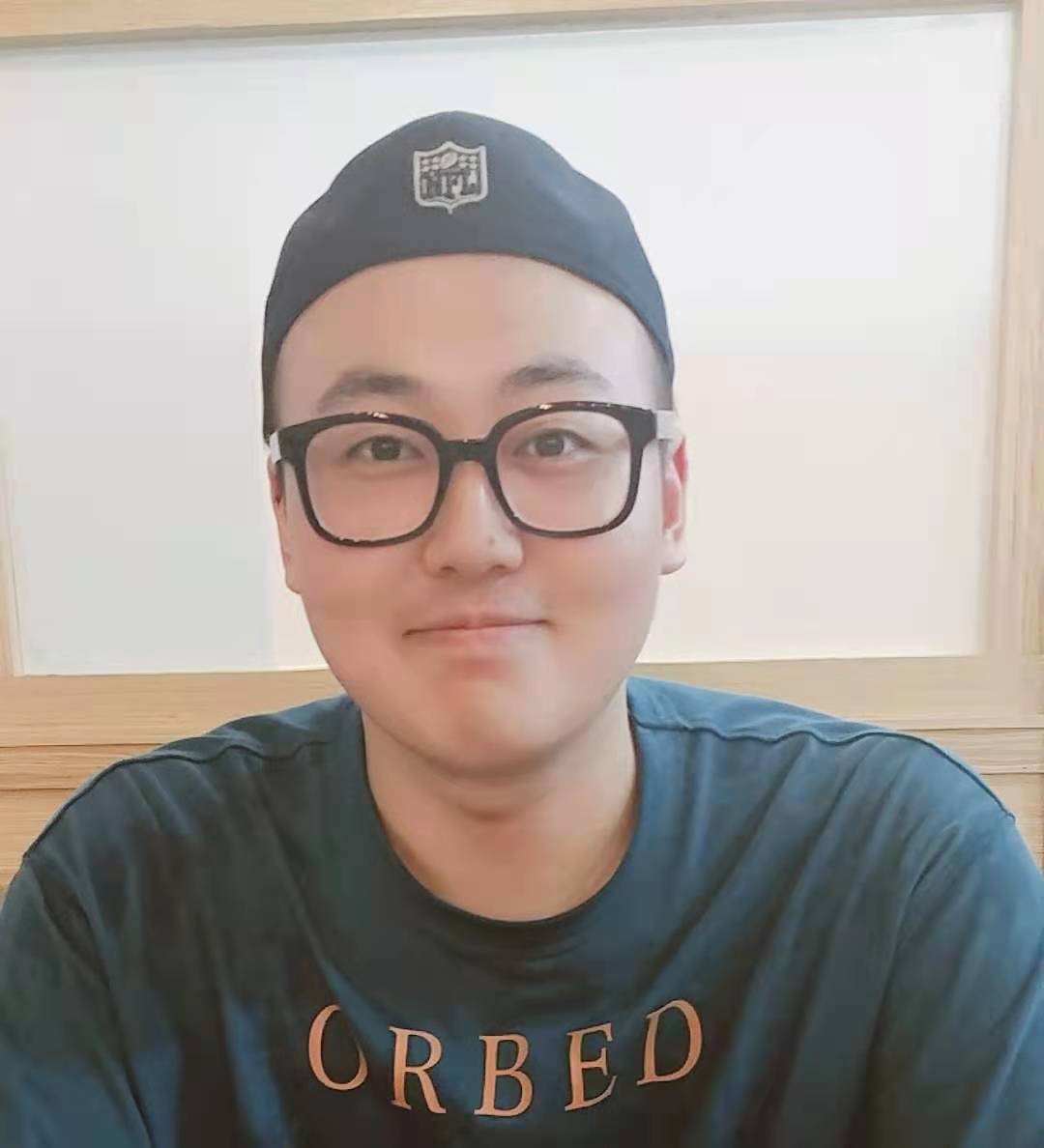}}]{}
\textbf{Zhenyu Zhang} is now an associate professor in Nanjing University. He received Ph.D. degree from Department of Computer Science and Engineering, Nanjing University of Science and Technology in 2020, supervised by Jian Yang. In 2019, he was a visiting student at MHUG group in University of Trento, Italy, supervised by Nicu Sebe. His research interests include 3D generation and perception, digital human.\end{IEEEbiography}

\begin{IEEEbiography}[{\includegraphics[width=1in,height=1.25in,clip,keepaspectratio]{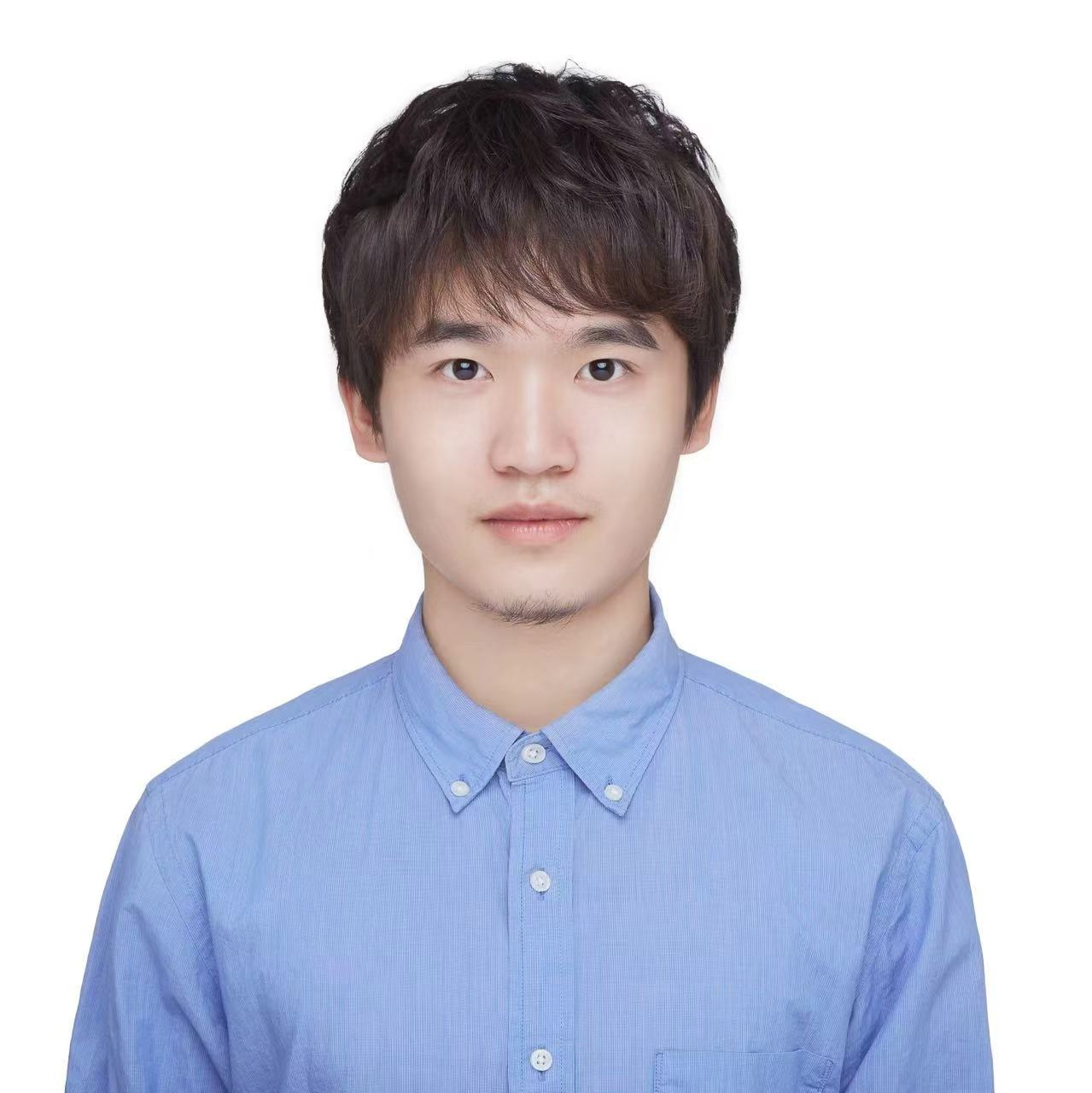}}]{}
    Yixuan Dong received the B.S. degree in Information and Computer Engineering from Chung Yuan Christian University, Taiwan, and the M.S. degree in Computer Science and Information Engineering from National Cheng Kung University, Taiwan, under the supervision of Prof. Jung-Hsien Chiang.
    He is currently a Researcher with the Center for Integrated Circuits and Artificial Intelligence, Tsientang Institute for Advanced Study. His research interests include computer vision, data augmentation, medical image analysis, and fairness and trustworthiness in artificial intelligence. His work focuses on developing robust and reliable machine learning methodologies for real-world and high-stakes applications, particularly in the medical and healthcare domains.
\end{IEEEbiography}

\begin{IEEEbiography}[{\includegraphics[width=1in,height=1.25in,clip,keepaspectratio]{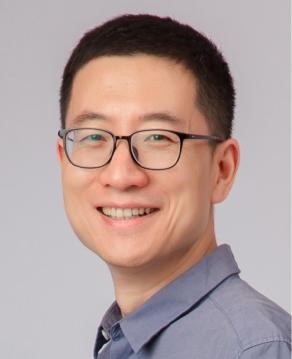}}]{}
\textbf{Jianjun Qian} (Member, IEEE) received the Ph.D. degree from Nanjing University of Science and Technology (NJUST), Nanjing, China, in 2014, with a focus on pattern recognition and intelligence systems.  He was selected as a Hong Kong Scholar in 2018.  He is currently a Processor with NJUST.  His research interests include pattern recognition theory, computer vision, and machine learning.  Dr. Qian has served as a Guest Editor for Neural Processing Letters and The Visual Computer.\end{IEEEbiography}

\begin{IEEEbiography}[{\includegraphics[width=1in,height=1.25in,clip,keepaspectratio]{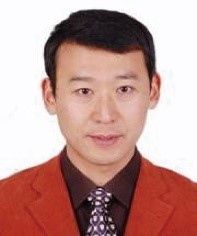}}]{}
\textbf{Jian Yang} received the PhD degree from Nanjing University of Science and Technology (NJUST) in 2002, majoring in pattern recognition and intelligence systems. From 2003 to 2007, he was a Postdoctoral Fellow at the University of Zaragoza, Hong Kong Polytechnic University and New Jersey Institute of Technology, respectively. From 2007 to present, he is a professor in the School of Computer Science and Technology of NUST. His papers have been cited over 60000 times in the Scholar Google. His research interests include pattern recognition and computer vision. Currently, he is/was an associate editor of Pattern Recognition, Pattern Recognition Letters, IEEE Trans. Neural Networks and Learning Systems, and Neurocomputing. He is a Fellow of IAPR.\end{IEEEbiography}

\vfill

\end{document}